\documentclass[11pt]{article}

\usepackage[utf8]{inputenc}
\usepackage[T1]{fontenc}
\usepackage[margin=1in]{geometry}
\usepackage{microtype}
\usepackage{booktabs}
\usepackage{graphicx}
\usepackage{tikz}
\usetikzlibrary{arrows.meta,positioning,decorations.pathreplacing}
\tikzset{
  svc/.style={draw, rounded corners, align=center, font=\small, inner sep=4pt, minimum height=10mm},
  flow/.style={draw, -{Latex[length=2mm]}},
}
\usepackage{amsmath}
\usepackage{xcolor}
\usepackage[numbers,sort&compress]{natbib}
\usepackage{hyperref}
\hypersetup{colorlinks=true,linkcolor=black,citecolor=black,urlcolor=blue}

\newcommand{\sysname}{Talk to TR}
\newcommand{\ceag}{cross-era analogical grounding}
\newcommand{\CEAG}{Cross-Era Analogical Grounding}
\newcommand{\ttft}{TTFT}

\title{The Living Library: Transforming Archival Collections into\\
Conversational Knowledge Systems\\[4pt]
\large Lessons from the Theodore Roosevelt Presidential Library}

\author{
Pengce Wang$^{1}$ \quad Lucia Ronchi Darre$^{1}$ \quad Matt Briney$^{2}$ \quad Michaell Bakalars$^{1}$ \\
Dan Rutkowski$^{1}$ \quad Ursula Hardy$^{1}$ \quad David Wolf$^{1}$ \quad Laura Hoffman$^{1}$ \\
Allen Kim$^{1}$ \quad Shawn Wright$^{1}$ \quad Juan Lavista Ferres$^{1}$ \\[3pt]
$^{1}$Microsoft \quad $^{2}$TRPL \\[2pt]
{\footnotesize\ttfamily \{pengcewang, lronchidarre, mbakalars, danrutkowski, hardyurs,}\\
{\footnotesize\ttfamily wolf.david, laura.hoffman, allen.kim, shwright, jlavista\}@microsoft.com \quad matt@trlibrary.com}
}
\date{}

\begin{document}
\maketitle

\begin{abstract}
We present the Living Library, an end-to-end framework for transforming fragmented digital archives into governed, conversational, in-person exhibit experiences. Developed and deployed at the Theodore Roosevelt Presidential Library, the framework comprises four layers: digitization and corpus creation, AI-powered processing, retrieval and reasoning, and an optional embodied conversational interface. The first three layers aggregate a 300,000-record collection, apply OCR and structured metadata enrichment for expert curatorial review, and publish records to a hybrid dense/semantic index. Expert review is conducted through the Archivist App, a curator-facing interface that supports correction of AI-generated transcriptions and metadata.
The governed corpus powers both a researcher-facing interface and Talk to TR, a continuously operating exhibit that embodies Theodore Roosevelt as a full-scale digital human within a museum environment. To support live, face-to-face interactions, Cross-Era Analogical Grounding reframes contemporary questions through documented historical parallels, allowing Roosevelt to address present-day topics without inventing facts. Dual-path retrieval and end-to-end streaming keep responses grounded and responsive. Layered watchdogs, visitor-session isolation, automated conversation management, and independently restartable services enable reliable unattended operation for hundreds of visitors. Avatar realism, spatial audio, lighting, staging, and conversational design are developed and evaluated as an integrated experience. Rather than report a controlled benchmark, we describe lessons from operating Talk to TR as a public exhibit and offer a transferable model for transforming archival collections into believable, in-person conversational experiences.
\end{abstract}

\section{Introduction}
\label{sec:intro}

Theodore Roosevelt's vast archival holdings, including letters, manuscripts, books, photographs, and other artifacts, sit
scattered across institutions and geographies, partially cataloged, and reachable
in practice only by specialists willing to travel to them and navigate complex archival catalogs. Digitization programs have worked on this for decades
\citep{terras2011rise,holley2009how}, but manual cataloging does not scale to
collections of hundreds of thousands of items, and a scanned page behind a search
box is still not the same thing as an accessible one: a visitor must know what to
search for, in a vocabulary the archive shares, before the archive will respond.
Museums and libraries increasingly seek to \emph{activate} their collections, to
let the public engage with a historical figure or era directly rather than read
static exhibit text, and to democratize access to primary sources otherwise
confined to the archive
\citep{cameron2007theorizing,parry2007recoding,tallon2008introduction}. This
paper asks how far that activation can go: whether a fragmented, partially
cataloged collection can be made not just searchable but conversable,
without sacrificing historical integrity.

We answer with the \emph{Living Library}: a replicable framework, developed and
deployed at the Theodore Roosevelt Presidential Library (TRPL), that layers (1)
digitization and corpus creation, (2) AI-powered processing, (3) a
retrieval-and-reasoning layer, and (4) an optional conversational interface on
top of an archival collection (\S\ref{sec:system}). The first three layers alone
already turn a fragmented collection into a unified, governed, semantically
searchable corpus, consumed today by a researcher-facing retrieval interface called \emph{Campfire} (TRPL's public-facing experience at \href{https://campfire.trlibrary.com/}{campfire.trlibrary.com})
(\S\ref{sec:system}); the fourth layer is what we explore in the greatest depth
in this paper, because it poses the hardest version of the underlying problem.
We built that fourth layer as \emph{\sysname{}}, rendering Theodore Roosevelt as a full-scale
digital human, on a large-format LED wall within a curated physical environment,
that converses with visitors in the first person. Crucially, \sysname{} is not merely a web chatbot with an avatar attached: it is a continuously operating
physical-digital exhibit engineered to sustain the illusion of a living
historical presence in a shared physical space. Embodying such a presence is
 unlike building a general-purpose chatbot or a fictional role-play
character: the subject is a real person whose statements can be verified
against the historical record, and the encounter is conducted face to face, in
real time, before a heterogeneous public audience.

Why build the fourth layer at all? Layers 1 to 3 already discharge the access problem as it is conventionally posed: they make the collection findable. But findability presumes a visitor who arrives with a question, in a vocabulary the archive recognizes, and with the patience to refine it. This describes a researcher, not a regular museum visitor. The audiences that presidential libraries and museums most want to reach are precisely those who do not yet know what to ask, and for whom a search box is not an invitation but a barrier of a different kind. The digital heritage literature has long argued that interpretation, not retrieval, is what moves a general audience \cite{cameron2007theorizing, parry2007recoding,tallon2008introduction}, and that visitors' engagement in a gallery is governed by framing, social context, and physical staging as much as by the information on offer \cite{vomlehn2001exhibiting, hornecker2006learning}. \textit{New Dimensions in Testimony} made the strongest version of this case: it is the experience of asking a survivor a question and being answered directly that carries the encounter, not the informational content of the answer \cite{artstein2014timeoffset, traum2015newdimensions}. Layer 4 is our attempt to extend that premise to a subject who left no recordings, where the answers cannot be pre-filmed and must instead be composed, in the moment, from the documentary record he did leave. The wager is that a person who would never search an archive will still ask a question of someone standing in front of them, and that the archive can be made to answer in that register without forfeiting its integrity. Whether the wager pays off in the terms that matter, is exactly what \S\ref{sec:limitations} flags as open.

Embodying a real historical figure imposes three requirements that are mutually in
tension. First, \textbf{faithfulness}: responses must reflect the historical record
and be attributable to primary sources rather than fabricated. Second,
\textbf{non-anachronism}: the figure must not reference people, events, or concepts
postdating his lifetime, nor adopt the tone of a modern AI assistant. Third,
\textbf{engagement under a latency budget}: a face-to-face avatar must begin
speaking with little perceptible delay and remain engaging on open-ended and
frequently contemporary questions that a century-old archive does not
anticipate.
Archival grounding supports faithfulness but introduces retrieval latency;
unconstrained generation supports engagement but admits fabrication and anachronism;
and the real-time budget constrains both. Deploying this as an unattended
public exhibit adds two further requirements: continuous
\textbf{autonomous operation} and believable \textbf{embodied presence}. These two requirements are
uncommon in conversational-AI benchmarks, yet they have been equally influential in shaping the architecture as the conversational tensions described above (\S\ref{sec:constraints}).

Traditional archival practice makes a historical figure's record difficult to activate for broad audiences: materials are fragmented across repositories; item-level cataloging is manual and inconsistent \citep{terras2011rise}; access still depends on travel, training, and institutional privilege despite digitization \citep{holley2009how,nguyen2020facilitating,muehlberger2019transkribus}; and backlogs leave much of the collection effectively unread. These constraints motivate our core question: \emph{how might institutions make vast, fragmented, and partially cataloged collections universally accessible, searchable, and interpretable, without sacrificing historical integrity?}

Our contributions are:

\begin{itemize}
  \item \textbf{The Living Library model} (\S\ref{sec:system}): a four-layer
  framework, including digitization/corpus creation, AI-powered processing, retrieval and
  reasoning, and an optional conversational interface, that turns a large
  primary-source collection into a governed, conversationally accessible corpus,
  deployed end to end at TRPL
  (Optical Character Recognition (OCR) and metadata extraction $\rightarrow$ object storage $\rightarrow$ hybrid dense/semantic index
  $\rightarrow$ retrieval-grounded persona). A curator-facing review application (the
  \emph{Archivist App}) keeps a human in the loop: it supports correction of AI-generated transcriptions and metadata and gives
  curators per-record publish/unpublish control over the shared index, without
  holding unreviewed records back from it (\S\ref{sec:ocr}).
  \item \textbf{\CEAG{}} (\S\ref{sec:ceag}), a retrieval-grounding strategy that
  reframes a contemporary query as a historically attested analog from the
  archive, enabling faithful engagement on modern topics.
  \item \textbf{Latency-aware dual-path retrieval} (\S\ref{sec:latency}) that
  combines synchronous self-routed retrieval with asynchronous prefetching, plus
  verbal-filler masking, inside a fully streaming automatic speech recognition
  (ASR) / large language model (LLM) / text-to-speech synthesis (TTS) / avatar
  rendering pipeline.
  \item \textbf{Museum-scale autonomous operation} (\S\ref{sec:autonomy}): a
  continuously running exhibit for unattended public deployment that sustains
  hundreds of visitors per day through automated conversation-lifecycle management,
  per-visitor session isolation, layered watchdogs with automatic fault recovery,
  and dynamic context management (\S\ref{sec:context}) that keeps a single all-day
  session warm within a bounded LLM context.
  \item \textbf{A non-blocking, fail-open safety stack} (\S\ref{sec:safety}) for a
  public kiosk, including three progressively heavier model-level guardrails (regex, an
  async small-model classifier, and a parallel output review) plus curator-gated
  deployment controls, engineered so that safety never stalls the avatar.
  \item \textbf{Physical-digital experience integration} (\S\ref{sec:presence}):
  \sysname{} is engineered as a spatial exhibit. It consists of a full-scale digital human on a
  large-format LED wall within a curated environment of staging, lighting, and
  spatial audio, all of which sustains a believable in-room historical presence rather than a
  screen-based interaction.
  \item \textbf{Design principles and a replicability framework}
  (\S\ref{sec:principles}, \S\ref{sec:replicability}): five governing principles
  and a six-step process by which another institution could apply the same model
  to its own collection, with or without a conversational-avatar layer.
  \item \textbf{A deployment experience report} (\S\ref{sec:eval}): first-token-latency
  measurements and qualitative deployment observations from running the exhibit in
  public, together with a candid discussion of the open questions that such a deployment surfaces but that a
  short study cannot settle (\S\ref{sec:limitations}).
\end{itemize}

\section{Background: Traditional Archival Practice}
\label{sec:background}

Before describing the Living Library, it is worth being precise about the
practice it departs from. Four features of traditional archival work recur
across institutions and motivate the framework below.

\paragraph{Fragmented collections.} Records from a single historical figure or a particular era are
 rarely held in one place: correspondence, photographs, publications, and
artifacts accumulate across libraries, historical societies, and private
holdings, often reconciled only informally. TRPL's own holdings span more than
forty repositories, acquired from both public
Library of Congress systems and authenticated institutional systems
(\S\ref{sec:system}). Fragmentation is the default, not the exception.

\paragraph{Manual cataloging limitations.} Cataloging at the item level is
time-intensive and, across a long institutional history,
might produce inconsistent metadata: different eras of cataloging apply different
vocabulary, granularity, and completeness standards to otherwise similar
material \citep{terras2011rise}.

\paragraph{Access constraints.} Engaging a collection in depth has traditionally
required physical access, specialized research training, or institutional
affiliation. These are exactly the barriers that digitization and, more recently,
AI-assisted access aim to lower
\citep{holley2009how,nguyen2020facilitating,muehlberger2019transkribus}.

\paragraph{Underutilized assets.} As a consequence of the three constraints
above, large portions of most collections remain unprocessed, undiscovered, or
simply unread, a standing loss of historical value that scales with collection
size rather than shrinking as digitization tooling improves, since backlogs grow
alongside acquisition.

\subsection{Core Problem}
These four features compose into a single question that motivates this paper:
\emph{how might institutions make vast, fragmented, and partially cataloged
collections universally accessible, searchable, and interpretable, without
sacrificing historical integrity?} The Living Library is our answer; Talk to TR extends that answer to the particularly demanding case of a face-to-face, real-time conversational interface, where responses must be generated on the spot and without human curation.

\section{Related Work}
\label{sec:related}

\paragraph{Persona and role-play language agents.}
A long line of work builds dialogue agents with a consistent persona, from
persona-conditioned generation \citep{zhang2018personachat} to LLMs that role-play
specific characters. \citet{shao2023characterllm} train an agent to become
a particular (often historical) figure from a constructed profile;
\citet{wang2023rolellm} benchmark and elicit role-playing ability at scale; and a
growing literature evaluates persona fidelity, for example through psychological
interviews \citep{wang2024incharacter}. These approaches largely bake the
character into model parameters or a static prompt. In contrast, \sysname{}
grounds the persona non-parametrically in a large corpus of the subject's
own primary sources, so its claims are attributable to real
documents rather than to a synthesized biography. This distinction matters because the subject is a real person whose record can be checked.

\paragraph{Interactive cultural-heritage and historical-figure installations.}
The closest prior systems make a real person conversationally interactive in a
public setting. \emph{New Dimensions in Testimony}
\citep{artstein2014timeoffset,traum2015newdimensions} lets visitors converse with
Holocaust survivors by retrieving pre-recorded human video answers; virtual
human interviewers such as SimSensei/Ellie \citep{devault2014simsensei} and the
broader embodied-conversational-agent tradition \citep{cassell2000embodied} pair
an animated agent with dialogue management. A defining property of these systems
is that they replay authored or recorded human utterances: this eliminates
fabrication but fixes the answer set, so genuinely open-ended or contemporary
questions fall outside coverage. Museum facing conversational systems are moreover
judged as much on framing, engagement, and interaction stability as on raw answer
quality \citep{vomlehn2001exhibiting,hornecker2006learning,sylaiou2022virtual,barbieri2023museum,casillo2022ontology},
within a broader digital heritage tradition
\citep{cameron2007theorizing,parry2007recoding,tallon2008introduction}. Unlike systems with a fixed answer set, \sysname{}
generates novel responses from a historical text archive, enabling open-ended and contemporary questions while preserving historical fidelity.

\paragraph{Retrieval-augmented and active retrieval.}
Retrieval-augmented generation grounds LLM output in an external corpus
\citep{lewis2020rag,guu2020realm,izacard2021fid,borgeaud2022retro}, typically using
dense or late-interaction retrievers \citep{karpukhin2020dpr,khattab2020colbert};
see \citet{gao2023ragsurvey} for a survey. Most
relevant to us is active retrieval, where the model decides whether
and when to retrieve: FLARE anticipates upcoming content and retrieves on
demand \citep{jiang2023flare}, and Self-RAG learns to retrieve and critique via
reflection tokens \citep{asai2023selfrag}. Our same-turn mechanism for deciding
whether the current user turn requires consulting the knowledge base is an
instance of this idea, but we deploy it under a hard real-time, embodied
constraint and extend it with (i) a second,
asynchronous path that prefetches the next turn's evidence and (ii)
verbal-filler masking of the synchronous retrieval gap, an efficiency angle that
active-retrieval work has not had to confront in a face-to-face setting.

\paragraph{Real-time, streaming, and embodied conversational agents.}
Reducing perceived latency in spoken interaction has a long history, from
incremental dialogue processing \citep{schlangen2011incremental} to recent
full-duplex speech foundation models \citep{defossez2024moshi}. On the embodiment
side, photorealistic talking-head synthesis
\citep{prajwal2020wav2lip,zhang2023sadtalker} drives avatar faces from speech.
\sysname{} differs from prior work by studying retrieval-grounded dialogue under
hard eal-time, face-to-face constraints and by reporting end-to-end
first-token latency (including grounding cost); we defer pipeline details to
\S\ref{sec:approach}.

\paragraph{Long-term memory in conversational agents.}
To persist information across turns and sessions, agents summarize and re-inject
memory. Generative Agents maintain a memory stream with reflection
\citep{park2023generative}, and MemGPT manages a bounded context like an operating
system's memory hierarchy \citep{packer2023memgpt}. \sysname{} uses short-horizon
memory to keep a visit coherent (per-turn state and per-visit summaries), but it
\emph{does not} maintain visitor-linked personal memory across separate visits.
Separately, the deployment keeps non-identifying operational logs for debugging
and evaluation under an explicit retention policy (\S\ref{sec:epdg}). This design
choice is motivated by an unusual deployment constraint: a single, long-lived
session serving a stream of anonymous visitors all day, which motivates our
watermark-based visitor isolation and in-place context refresh
(\S\ref{sec:context}).

\paragraph{Historical document digitization and archive access.}
Making historical collections machine-readable relies on OCR and handwritten-text
recognition \citep{muehlberger2019transkribus,smith2007tesseract}, whose accuracy
is judged against downstream use rather than transcription alone
\citep{holley2009how,nguyen2020facilitating,terras2011rise} and whose residual
errors propagate into any downstream index. \sysname{} sits downstream of such
digitization: its corpus is built by a governed OCR and enrichment pipeline in
which handwritten material remains the dominant error source, and, rather than a search interface, it exposes the
reviewed collection through a grounded conversational persona.

\paragraph{Faithfulness, attribution, safety, and guardrails.}
Hallucination and its measurement are well studied
\citep{ji2023hallucination}, as is attribution to identified sources
\citep{rashkin2021ais}; retrieval grounding is known to reduce conversational
hallucination \citep{shuster2021retrieval}, which our design leans on directly. For public deployment, LLM applications must also resist
prompt injection \citep{greshake2023injection} and enforce content policy via
guardrails \citep{rebedea2023nemo,inan2023llamaguard}. We combine these concerns
for a public kiosk with a layered, fail-open safety stack that never blocks the
avatar, and we treat anachronism (references postdating the figure's
lifetime) as a distinctive faithfulness concern for historical-persona agents
(\S\ref{sec:eval}).

\section{The Living Library Model}
\label{sec:system}

\subsection{Architecture Overview}

The Living Library couples an offline corpus-construction pipeline with a set
of real-time services that consume the resulting governed index. In the TRPL
exhibit, the production path is intentionally decomposed into independently
restartable services with distinct failure modes: sensor fusion (presence and
visitor lifecycle), conversation orchestration (grounded reasoning over the
constructed corpus), a real-time agent bridge (streaming ASR/TTS and interruption
handling), and a rendering/presentation stack that delivers the avatar and
physical A/V output.

\begin{figure}[t]
\centering
\begin{tikzpicture}[
  lyr/.style={draw, rounded corners, align=left, font=\small,
              text width=0.80\linewidth, inner sep=5pt},
  node distance=5mm]
  \node[lyr, fill=blue!3] (l1)
    {\textbf{Layer 1: Digitization \& Corpus Creation}\\
     \footnotesize TRPL's digital collection $\rightarrow$ \textbf{Azure Storage}
     preservation copies (source IDs + rights preserved)};
  \node[lyr, fill=blue!7, above=of l1] (l2)
    {\textbf{Layer 2: AI-Powered Processing}\\
     \footnotesize \textbf{Azure AI Foundry}: OCR + metadata enrichment $\rightarrow$
     \textbf{Azure Cosmos DB} governed store $\rightarrow$ \textbf{expert curator review}};
  \node[lyr, fill=blue!11, above=of l2] (l3)
    {\textbf{Layer 3: Retrieval \& Reasoning}\\
     \footnotesize \textbf{Azure AI Search} hybrid index; cross-era analogical grounding + dual-path retrieval
     \quad$\rightarrow$\quad text interface: \textbf{Campfire}};
  \node[lyr, fill=blue!15, above=of l3] (l4)
    {\textbf{Layer 4: Embodied Conversational Interface} \emph{(optional)}\\
     \footnotesize \sysname{}: voice + avatar + physical exhibit (adds embodiment over Layer 3)};
  \draw[flow] (l1) -- (l2);
  \draw[flow] (l2) -- (l3);
  \draw[flow] (l3) -- (l4);
\end{tikzpicture}
\caption{The four-layer \textit{Living Library} model, deployed at TRPL (data flows
upward). Layers 1 and 2 turn fragmented sources into a governed, searchable corpus
through OCR, metadata enrichment, and expert review; Layer 3 grounds generation in that
corpus; and the optional Layer 4 adds a conversational interface, 
\textit{\sysname{}}. Layers 1 to 3 alone already serve the researcher-facing \textit{Campfire}.}
\label{fig:kb}
\end{figure}
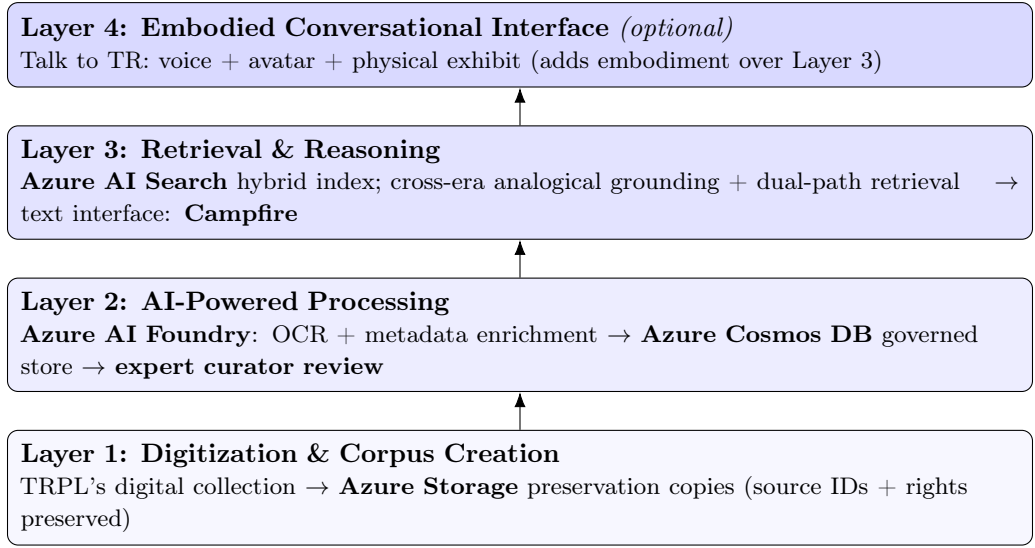

Figure~\ref{fig:kb} shows the cloud pipeline behind the corpus. Material is
extracted from a private API provided by TRPL into \textbf{Azure Storage}, which holds
institution-controlled preservation copies. Each page image then passes through
\textbf{Azure AI Foundry}, where LLMs perform OCR and structured metadata extraction
(\S\ref{sec:ocr}); the resulting text, per-item confidence, and soft-metadata fields
are written to \textbf{Azure Cosmos DB} alongside preserved hard metadata, which
defines the record schema and serves as the system of record. Records are chunked, embedded with an embedding model, and
published to \textbf{Azure AI Search}. The index's schema and semantic configuration
govern retrieval at query time for downstream retrieval-grounded interfaces (the
\sysname{} avatar and the researcher-facing Campfire). This keeps retrieval aligned
with the governed corpus contract (\S\ref{sec:principles}). The index is also consumed by the Archivist App, which enables expert
curators from TRPL to verify transcriptions and metadata enrichment against the
source images, making necessary edits whenever needed. These reviews feed back into the index, allowing for continuous improvement of the data without holding back items that have not been reviewed yet. 

\subsection{Layers 1 and 2: Digitization, Corpus Creation, and AI-Powered Processing}
\label{sec:layers12}
\sysname{} draws on TRPL's holdings, roughly
300{,}000 records. These include major, public collections such those from the Library of Congress and Harvard College Library, as well as other smaller, private ones. 

\textbf{Layer 1 (digitization and corpus creation)} ingests these sources while preserving source identifiers and rights status into
institution-controlled preservation storage, so downstream stages never depend on
fragile upstream paths. 

\textbf{Layer 2 (AI-powered processing)} then passes the images as inputs to an LLM, generating either text transcriptions (for text records, such as letters) or visual descriptions (for visual records, such as photographs). The original metadata (such as Collection name) remains immutable and is separate from AI-generated values. The data is then published to a search index, which feeds into the Archivist App, a web interface developed specifically for TRPL's archivists to review and edit AI-generated values. After expert curator
review, the records are  re-published into the index and marked as reviewed.

Independently, Layers 1 and 2 already turn the fragmented
collection into a unified, searchable resource for researchers, before any
conversational layer is added (\S\ref{sec:replicability}).

\subsection{Layer 3: Retrieval and Reasoning}
\label{sec:layer3}
Layer 3 grounds generation in the verified corpus rather than the model's
parametric memory: queries are interpreted, relevant material is retrieved from the index, and an LLM composes a grounded,
attributable response. This is what powers the digital experience (Campfire). For the in-person experience (Talk to TR), retrieval is further constrained by two additional mechanisms detailed 
in \S\ref{sec:approach}. The first mechanism is the \CEAG{} (\S\ref{sec:ceag}), which reframes contemporary
queries as historically attested analogs so the archive can answer questions it
never anticipated. The second mechanism is a latency-aware dual-path retrieval (\S\ref{sec:latency}),
which keeps that grounding affordable inside a real-time conversational budget.

Layer 3 is where the framework's general claim is made concrete and testable; grounding in a governed
corpus keeps synthesis attributable. Exposed through a
web experience with no embodiment, Layer 3 is already usable on its own: this is
exactly TRPL's researcher-facing \emph{Campfire}.

\subsection{Layer 4: Embodied Conversational Interface,  \sysname{}}
\label{sec:layer4}
Layer 4 is optional and additive. Layers 1 to 3 are already usable on their own as a
text-driven conversational interface: TRPL's is \emph{Campfire}, a chat over the
Layer-3 corpus with no embodiment. What Layer 4 adds is embodiment: voice,
avatar, and physical presence over the same corpus. TRPL's instantiation is
\sysname{}, a full-scale digital human delivered as a continuously-operating
physical-digital museum exhibit rather than a web chatbot.

The avatar's in-scope knowledge base is a curator-reviewed subset of this corpus
(Roosevelt's writings, letters, speeches, biographies, and approved interpretive
content), retrieved through the same governed access contract as TRPL's
researcher-facing interface, \emph{Campfire}.

\paragraph{Interaction loop.}
A visitor ``conversation'' is a long-lived session driven by a lightweight phase
state machine. In production the exhibit uses two core phases: an \textbf{idle}
phase (ambient storytelling and attract behavior) and a
\textbf{storytelling} phase (the grounded question-answering loop), with
transitions triggered by the sensor-fusion aggregator (a visitor stepping up to
the lectern) and by operator/RFID signals. Ending a conversation returns the
session to the idle phase rather than terminating it, so that the next visitor
incurs no cold-start cost (\S\ref{sec:context}).

\paragraph{Turn ownership and stale-generation containment.}
Each accepted visitor question is assigned a generation (request identity)
that follows it through transcription, the orchestrator, retrieval, TTS, and playback. When a
visitor deliberately begins a new question, ownership moves to the new generation
and all older in-flight results are quarantined, even if an LLM, retrieval request,
or TTS operation completes later. Interruptions trigger generation-scoped audio
and caption disposal: the Unreal Engine ``show'' that renders and composites
the avatar for the physical exhibit fades out the active voice, clears stale
procedural audio, rejects delayed resume events from older generations, and fades in
only the newest valid generation. This prevents mixed answers, stale captions, delayed audio
from a prior question, and ``Frankensteined'' responses assembled from two turns.

\paragraph{Rendering service.}
Generated text is streamed to \textbf{Azure Custom Voice}, trained on a voice actor's
studio recordings for in-character vocal continuity, and to Lemon Slice, a real-time
GenAI avatar provider that synchronizes speech, facial expression, and gesture.
LiveKit carries the real-time media session.

For the physical exhibit, a packaged \textbf{Unreal Engine} ``show'' consumes LiveKit media
through a custom native plugin that uploads incoming video frames directly into a
runtime texture and performs final composition, synchronization, and presentation.
Because cloud frames can arrive at varying resolutions, the show includes a GPU
upscaling path (TensorRT Real-ESRGAN) with frame holding/interpolation for stable
LED-wall output and continuous A/V drift measurement for persistent delay
correction. A stencil-composite post-process excludes the avatar texture from
Unreal's normal temporal reconstruction history to prevent facial ghosting.

Audio is likewise treated as exhibit infrastructure: a custom Unreal audio module
claims the Dante Virtual Soundcard---a networked digital-audio interface---through the
low-latency ASIO (Audio Stream Input/Output) driver, transports samples over a
lock-free single-producer/single-consumer ring from the audio-render thread to the
ASIO callback, and exposes stem routing (voice/music/ambience/effects) for a
multi-speaker installation.

The exhibit also treats captions and operator control as first-class integration
surfaces. Transcript updates carry message IDs and monotonic sequence numbers for
deduplication and grouping; responses include estimated speech-duration
metadata to support timed lectern display. Unreal publishes transcript state via
its Remote Control WebSocket interface, and a separate live-caption path emits
progressively accumulated text paced to synthesized speech so captions appear
alongside delivery rather than as a single block after playback.

\subsection{Embodied Physical Presence}
\label{sec:presence}
The deployment environment extends well beyond the conversational system. Responses
are delivered through a full-scale digital human rendered on a large-format LED wall
and integrated into a curated physical environment, including architectural staging,
environmental lighting, spatial audio, and deliberate visitor positioning. The
objective is not information retrieval but a believable in-room encounter with a
historical figure; the blending of physical and digital presentation is what
sustains the visitor's willing suspension of disbelief. This requirement shaped
rendering, audio, interaction design, visitor flow, and the operational architecture
alike.

\subsection{Deployment Context and Observations}
\label{sec:constraints}

Public-facing conversational exhibits impose requirements uncommon in conventional
conversational-AI benchmarks. The system must operate continuously throughout
exhibition hours, tolerate unpredictable visitor behavior, support rapid visitor
turnover, recover gracefully from infrastructure failures, and sustain engagement
without technical supervision. It must, moreover, deliver its responses not on a
screen the visitor holds but as a life-sized presence in a shared physical space
(\S\ref{sec:presence}). The requirements on continuous
autonomous operation (\S\ref{sec:autonomy}) and believable embodied
presence (\S\ref{sec:presence}) shaped the architecture of \sysname{} as strongly
as the conversational requirements of faithfulness, non-anachronism, and real-time
engagement (\S\ref{sec:intro}), and they are the respects in which a deployed exhibit
most departs from a research prototype evaluated in short sessions.

\paragraph{Deployment observations.}
\label{sec:eval}
We report what a live public exhibit lets us measure, and are explicit about what it
does not. \sysname{} has run in front of real visitors rather than in a controlled
study: in its first two weeks of public operation (July 2026) it engaged close to
5{,}000 visitors, so our evidence is largely observational, combining basic usage
logging with qualitative deployment observation. Running the exhibit made clear what actually
governs a good visit. Operationally, we watched whether turns complete, how often the
system falls back to a safe response, and whether the persona stays within its
approved bounds, alongside curator judgments of grounding and historical restraint.
Experientially, the strongest lesson was that immersion depended on far more than the
AI: avatar realism, display fidelity, lighting, audio, and physical staging all
mattered, and small changes to the environment could visibly shift the sense of
presence with no change to the underlying system. A historical digital human of this
kind is therefore best judged as a complete experiential system, not a language
interface alone. By extension, the Living Library framework that produces it is
best judged by what it enables end to end, not by any one layer in isolation.

\section{Technical Architecture}
\label{sec:approach}

\begin{figure}[t]
\centering
\includegraphics[width=\linewidth]{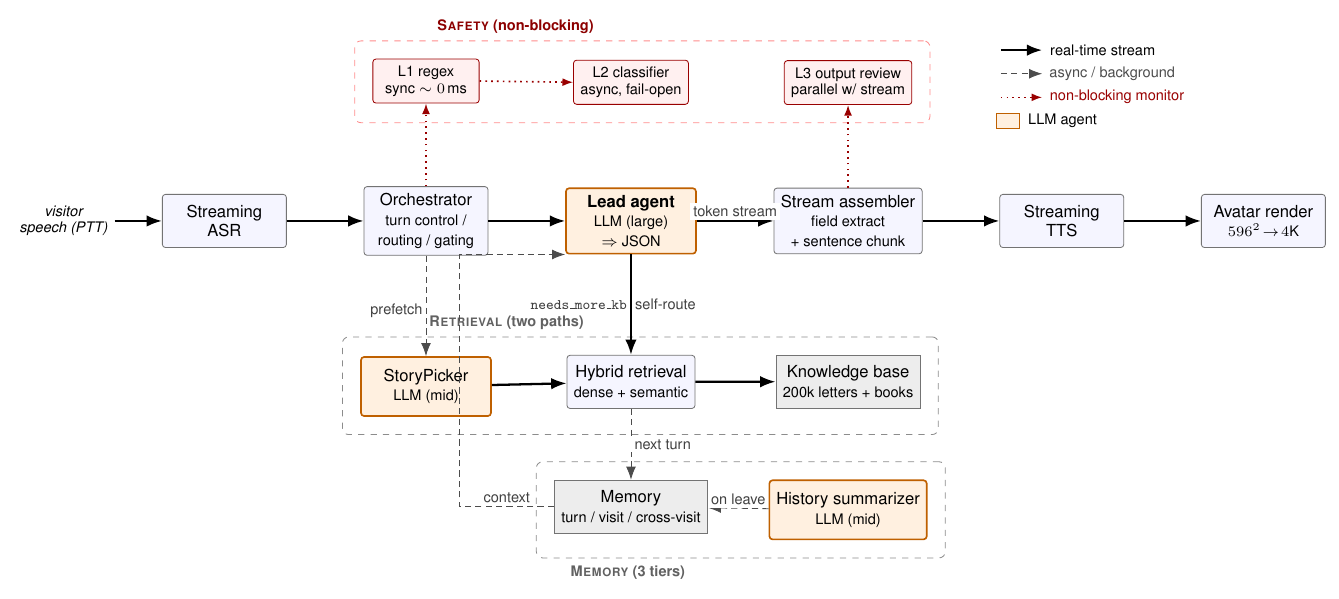}
\caption{Internal architecture of the dialogue service. The real-time streaming
path (bold) runs from visitor speech through streaming automatic speech recognition (ASR), the orchestrator, the
\textbf{lead agent}, a stream assembler, streaming TTS, and the avatar renderer.
The LLM agents (orange)---the lead speaker, StoryPicker, and history
summarizer---are coordinated around two retrieval paths (asynchronous
\emph{prefetch} and synchronous \emph{self-routing} on \texttt{needs\_more\_kb};
\S\ref{sec:latency}), a three-tier memory (\S\ref{sec:context}), and a
non-blocking, three-layer safety stack (\S\ref{sec:safety}). Solid arrows denote
the real-time stream; dashed, asynchronous/background flow; dotted, non-blocking
monitors.}
\label{fig:brain}
\end{figure}

Having situated \sysname{} within the Living Library model and its deployment
context, we now detail the technical steps involved, including OCR construction
and model selection, persona and voice, cross-era analogical grounding,
latency-aware retrieval, dynamic context management, safety, and autonomous
operation. Figure~\ref{fig:brain} shows how the dialogue-service elements are
organized.

\subsection{OCR Construction, Model Selection, and Curatorial Review}
\label{sec:ocr}
Because retrieval quality is bounded by transcription quality, the OCR stage is
constructed and evaluated explicitly rather than adopted off the shelf. For this exercise, records were obtained from the Library of Congress API, which contains text transcriptions for a portion of the records in the Theodore Roosevelt Papers collection. The OCR stage itself is model-pluggable:
each candidate runs as an interchangeable module writing model-specific text,
per-item confidence, and metrics alongside the record, so the production model is a
configuration choice rather than a hard-wired dependency (\S\ref{sec:principles},
\emph{scalable by design}). Evaluated candidates span Mistral OCR, DeepSeek, GPT-4o,
GPT-4.1, and GPT-5.

\paragraph{Source-grounded evaluation.}
Model choice is fixed by a source-grounded evaluation rather than ad hoc inspection.
We draw a stratified sample across production methods so that
handwritten and typed material are both represented, and score each model's output against reference
text. Each output is scored on character and word
error rate \citep{rice1996measuring} and, because the downstream task is
retrieval-grounded dialogue rather than literal transcription, also on sentence-level
BLEU \citep{papineni2002bleu} and cosine similarity of
text embeddings \citep{reimers2019sentencebert}. Including
a semantic axis is deliberate: OCR output that is lexically imperfect but
semantically faithful can still support retrieval, which is what ultimately matters
downstream \citep{nguyen2020facilitating}.

\begin{table}[t]
\centering
\begin{tabular}{lcccc}
\toprule
OCR model & Mean CER$\downarrow$ & Mean WER$\downarrow$ & BLEU$\uparrow$ & Emb.\ cos.$\uparrow$ \\
\midrule
GPT-5             & \textbf{0.124} & \textbf{0.098} & \textbf{0.861} & \textbf{0.948} \\
GPT-4.1 (default) & 0.130 & 0.099 & 0.852 & 0.944 \\
GPT-4o            & 0.154 & 0.141 & 0.816 & 0.942 \\
Mistral OCR       & 0.154 & 0.143 & 0.811 & 0.913 \\
DeepSeek          & 0.220 & 0.174 & 0.786 & 0.880 \\
\bottomrule
\end{tabular}
\caption{OCR model comparison on a grounded, stratified validation sample.
Lower CER/WER and higher BLEU/embedding-cosine are better;
best per column in \textbf{bold}. GPT-5 leads on every metric, while GPT-4.1 (the
production default) trails only narrowly.}
\label{tab:ocr}
\end{table}

\paragraph{Production decision.}
Across the comparison (Table~\ref{tab:ocr}), GPT-5 leads on every metric, while GPT-4.1
trails it only narrowly and, at deployment time, offered the best balance of quality,
cost, and throughput under batch processing; it is therefore the production default.
The semantic axis matters here: the strongest models cluster tightly on embedding
cosine (all ${\geq}\,0.94$) even where character error differs, confirming their
outputs are retrieval-ready. Handwritten material remained the dominant error source
across models, which is why expert review
(\S\ref{sec:system}) is important.

\paragraph{The Archivist App: human review of AI output.}
Processed records are routed for human review through the \emph{Archivist
App}, a curator-facing web application built for TRPL's archivists. Curators
search and filter the corpus by source, resource type, review status, and OCR
confidence, open a record beside its source page image, and correct the
AI-generated transcription and soft-metadata fields in place. Edits are non-destructive and versioned: the original model output is preserved
immutably, and each correction is written as a new version of a separate
transcription, with a side-by-side original-versus-edited comparison and a full
audit trail recording who changed what, when, and across which version
transition. Hard metadata carried from the source system (for example, collection
provenance and rights) remains immutable throughout, so human correction never
overwrites the institutional system of record.

\paragraph{Continuous indexing with curatorial correction and withdrawal.}
Review is not an admission gate: processed records are published to
the search index continuously, so unreviewed material is not held back from downstream retrieval.
What the Archivist App adds is curatorial control over that index rather
than a precondition for entering it. Curators correct and verify records in place,
and their edits are re-published so the corpus improves continuously; per-record
publish and unpublish controls let a curator push a corrected record into the
index or withdraw a problematic one from downstream retrieval at any time. This
reconciles curatorial accountability with a 300{,}000-record backlog: the
collection becomes searchable immediately, while human review raises its quality
over time instead of blocking it.

\subsection{Production Service Topology and Fault Containment}
\label{sec:topology}
The physical exhibit is engineered as a set of small services whose boundaries match the
real operational failure modes of a museum installation. The end-to-end
production path is:
\emph{camera/RFID/pressure plate} $\rightarrow$ \emph{sensor-fusion aggregator}
$\rightarrow$ \emph{conversation orchestrator} $\rightarrow$ \emph{real-time agent
bridge} $\rightarrow$ \emph{Azure voice and Lemon Slice} $\rightarrow$ \emph{LiveKit}
$\rightarrow$ \emph{custom Unreal Engine plugin} $\rightarrow$ \emph{LED wall and
physical audio system}.

The aggregator produces an authoritative visitor lifecycle (who is present, who
``owns'' the session, and which interaction mode is active) by arbitrating camera
zones, RFID identity, and pressure-plate occupancy: it debounces movement, clears
anonymous slots when the lectern empties, and prevents stale camera tracks from
blocking the next visitor. The orchestrator governs grounded conversation and archival retrieval. The agent
bridge manages streaming speech recognition and synthesis, fillers, interruptions,
and the avatar session. Unreal performs final video composition, audio
routing/synchronization, and physical presentation.

\begin{table}[t]
\centering
\begin{tabular}{p{0.27\linewidth}p{0.68\linewidth}}
\toprule
\textbf{Layer} & \textbf{Detection and recovery} \\
\midrule
Unreal Engine & Process/game-thread heartbeat; automatic show relaunch on renderer crash or stall.\\
Orchestrator & HTTP health plus a one-token provider probe; restart when the process is alive but the graph is unresponsive.\\
Agent bridge & Process, health-socket, and turn-progress monitoring; bridge-only restart while Unreal remains running.\\
Aggregator & Process and sensor-freshness monitoring; restart when camera/RFID delivery becomes stale.\\
Lemon Slice & Video-FPS, returned-audio, first-frame, and participant-state checks; rebuild the cloud session when media becomes unusable.\\
ASIO/Dante & Mandatory startup claim, callback/ring-buffer health telemetry, and prevention of silent fallback to Windows audio.\\
ISAAC control bridge & Independently supervised museum-control endpoint enabling operator-triggered recovery when automatic healing is insufficient.\\
\bottomrule
\end{tabular}
\caption{Layered fault containment and recovery. The design principle is escalation
from the least disruptive recovery (rebuild the avatar session) to more disruptive
steps (restart the agent, restart the orchestrator/aggregator) and only then a full show
restart.}
\label{tab:recovery}
\end{table}

\subsection{Grounding a Historical Persona}
\label{sec:persona}

The persona is defined by a layered system prompt built once and held resident.
The innermost layer is a \emph{persona core} specifying the subject's voice,
values, and diction, together with \emph{governance rules}: an avoid-list of
modern words and topics, instructions for handling sensitive or contested history,
factual-honesty constraints, and prompt-injection resistance. Around it, a
\emph{behavior} layer fixes the scene, turn management, and hand-off conventions;
an \emph{audience} layer selects an adult or child variant; and an
\emph{interface} layer specifies the input/output contract. A small number of
\emph{few-shot exemplar exchanges} constrain tone, sentence length, and rhythm more
reliably than abstract style rules, and explicit length and question-frequency
rules---a short reply to a short input, expansion only for open questions, and the
avoidance of unsolicited exposition---prevent the persona from drifting into a
verbose assistant register.

\paragraph{Structured turn output.}
Rather than issue separate calls for \emph{what to say} and \emph{how to manage the
turn}, the speaker LLM emits one structured (JSON) object per turn that carries both
the spoken line and the control signals the orchestrator needs:
\[
  \{\,\texttt{response},\ \texttt{target},\ \texttt{memory},\ \texttt{done},\ \texttt{needs\_more\_kb}\,\}.
\]
Only \texttt{response} is voiced. The remaining fields are control signals:
\texttt{target} (whom the reply addresses), \texttt{memory} (facts to persist across
the session, \S\ref{sec:context}), \texttt{done} (end-of-turn), and
\texttt{needs\_more\_kb} (the same-turn retrieval-routing decision of
\S\ref{sec:latency}). Folding speech and control into a single call is a latency
choice: the orchestrator extracts and begins streaming \texttt{response} to TTS by
incremental JSON parsing \emph{while} the trailing control fields are still arriving,
so each turn costs one LLM round-trip rather than two.

Because on-site tuning of tone and policy is frequent, the persona prompts are
\emph{overridable at runtime} through a small keyed store, so a curator can adjust
the voice without code changes or redeployment (\S\ref{sec:principles}); overrides
persist to a file and take effect for subsequently built sessions.

\subsection{Cross-Era Analogical Grounding}
\label{sec:ceag}

A century-old archive cannot directly answer a question about a contemporary
subject such as social media or electric cars. Unconstrained generation risks
anachronism and fabrication, whereas refusal is unengaging. \CEAG{} instead
reframes a contemporary query as a retrieval for a \emph{historically attested
analog}.

\begin{figure}[t]
\centering
\begin{tikzpicture}[node distance=6mm and 10mm]
  \node[svc] (q) {query $q$\\(often modern)};
  \node[svc, below=of q] (pick) {Picker (mid LLM)\\$\to\langle t,s,k,\rho\rangle$};
  \node[svc, right=of pick] (cat) {108-story catalog\\(resident in prompt)};
  \node[svc, below=of pick] (retr) {retrieve $\mathcal{E}{=}\mathrm{Retr}(k)$\\+ load story $s$};
  \node[svc, below=of retr] (llm) {Speaker LLM\\$r{=}\mathrm{LLM}(q,s,\mathcal{E},\rho)$};
  \draw[flow] (q) -- (pick);
  \draw[flow,dashed] (cat) -- (pick);
  \draw[flow] (pick) -- node[right,font=\scriptsize]{$\star$ modern$\to$era analog} (retr);
  \draw[flow] (retr) -- (llm);
\end{tikzpicture}
\caption{\CEAG{}: a contemporary query is reframed as retrieval for a
historically attested analog, then answered in period and in voice.}
\label{fig:ceag}
\end{figure}
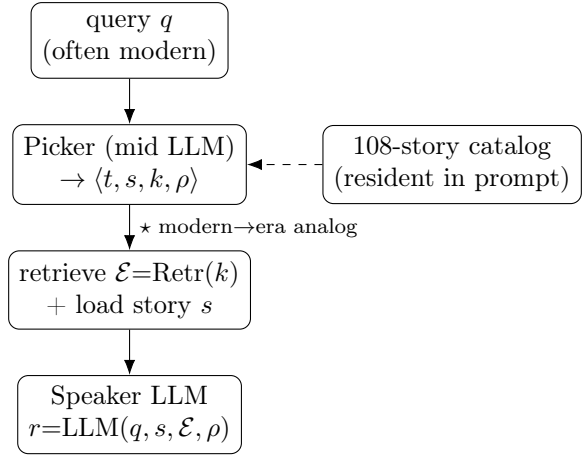

Given a visitor query $q$, a \emph{picker} module (a mid-tier LLM) produces a
tuple $\langle t, s, k, \rho\rangle$: an era-appropriate target theme $t$, a
selected story $s$ from a resident catalog of 108 curated stories, a retrieval
query $k$ over the archive, and a short rationale $\rho$ explaining why $s$ is
relevant to $q$. We then retrieve evidence
$\mathcal{E} = \mathrm{Retr}(k)$---letters and book excerpts---via hybrid
dense/semantic search (Layer 3, \S\ref{sec:layer3}), load the full narrative of $s$, and have the speaker LLM
generate
\[
r = \mathrm{LLM}(q,s,\mathcal{E},\rho,\mathrm{persona}),
\]
answering $q$ \emph{through} the analog while remaining in-voice and in-period.
The catalog of story titles and one-line hooks is held resident in the system
prompt, so that the model is aware of the available stories without a retrieval
call; the rationale $\rho$ (a ``curator hint'') enables the speaker to integrate the
story rather than recite it. A per-turn and cross-turn memory of stories already
told enforces rotation (no repeats), and a story is unloaded from context after a
bounded number of turns to admit a fresh one. Calendar-anchored ``this day in
history'' facts are injected similarly, enabling the figure to introduce timely
material.
Figure~\ref{fig:ceag} summarizes the flow.

\paragraph{Worked example.}
Figure~\ref{fig:ceag-example} traces a single turn end to end. Asked ``What do you
think about social media?''---a subject with no direct archival referent---the picker
does \emph{not} answer it directly. It reframes the query into an era-appropriate
theme (reaching the people directly), selects a catalog story about how Roosevelt
moved the public with his own voice and the press, emits a retrieval query, and
attaches a one-line rationale linking the modern subject to the analog. Retrieval then
returns real archive evidence---here, passages on Roosevelt's use of the presidency as
a ``bully pulpit''---and the speaker composes an answer that engages the modern topic
\emph{through} that analog, in period and in voice, without naming any technology
that postdates the figure's life.

\begin{figure}[t]
\centering
\small
\begin{tabular}{@{}p{2.2cm}p{10.5cm}@{}}
\toprule
\multicolumn{2}{@{}p{12.7cm}@{}}{\textbf{Visitor query} $q$:\; ``What do you think about social media?''}\\
\midrule
\multicolumn{2}{@{}p{12.7cm}@{}}{\textit{Picker} (mid-tier LLM) $\;\Rightarrow\;\langle t,s,k,\rho\rangle$:}\\
theme $t$   & reaching the people directly, over the gatekeepers of the day \\
story $s$   & \textit{``Words Sharper Than Swords''}---how Roosevelt moved the public with his voice and pen (from the resident story catalog) \\
query $k$   & ``Roosevelt \;/\; the press \;/\; the `bully pulpit' \;/\; appealing directly to the people'' \\
hint $\rho$ & social media lets one address the public directly; Roosevelt's analog is the mass-circulation press and the ``bully pulpit'' \\
\midrule
evidence $\mathcal{E}$ & retrieved excerpts from \textit{The Bully Pulpit} and Roosevelt's own writing on using the presidency as a ``bully pulpit'' to reach citizens over the party bosses \\
\midrule
\multicolumn{2}{@{}p{12.7cm}@{}}{\textbf{Answer} $r{=}\mathrm{LLM}(q,s,\mathcal{E},\rho,\text{persona})$, voiced in period:\; ``Social media, you call it? In my day the newspapers were my bully pulpit---I learned to speak straight to the people, over the heads of the party bosses. Give a citizen a just cause and the means to be heard, and he can rouse the whole country.''}\\
\bottomrule
\end{tabular}
\caption{A representative \CEAG{} interaction, end to end. Story and evidence are drawn
from the deployed catalog and corpus; the spoken answer is abridged. The modern
subject is engaged \emph{entirely} through a historically attested analog, with no
post-1919 reference.}
\label{fig:ceag-example}
\end{figure}

By design, \CEAG{} sits between two simpler alternatives---free role-play with no
retrieval, which maximizes engagement but admits anachronism and fabrication, and
retrieval without analogical reframing, which grounds answers but struggles to bridge
a modern query to a century-old record. Reframing is what lets the figure engage a
modern topic while staying attributable and in period; \S\ref{sec:limitations} discusses the
open questions this design raises.

\subsection{Meeting the Real-Time Budget}
\label{sec:latency}

Grounding a face-to-face avatar in an archive is only useful if it stays
responsive. We optimize \emph{first-token latency} (\ttft{}), the delay from
end-of-question to the first spoken word (the utterance then continues to stream),
since a face-to-face encounter demands a prompt reply. Three design choices keep
responsiveness acceptable under this budget.

First, \textbf{local streaming automatic speech recognition (ASR)}: while a visitor holds the push-to-talk button,
a local Whisper \texttt{large-v3} decoder advances every $0.15$\,s over a rolling
window, maintaining a $\sim$4\,s correction horizon that can revise recent words
while committing older, stable text during long questions. On button release, a
revision-aware flush processes only the remaining audio tail rather than
retranscribing the entire recording.

Second, a \emph{dual-path} retrieval scheme: the speaker self-routes with a
\texttt{needs\_more\_kb} flag and, when set, retrieves \emph{in the current turn}
and regenerates \citep{jiang2023flare,asai2023selfrag}, while an asynchronous path
prefetches the next turn's evidence as the current answer streams.

\begin{figure}[t]
\centering
\begin{tikzpicture}[node distance=4mm and 7mm]
  \node[svc] (asr) {ASR};
  \node[svc, right=of asr] (llm) {LLM};
  \node[svc, right=of llm] (tts) {TTS};
  \node[svc, right=of tts] (av) {avatar\\frames};
  \node[svc, right=of av] (up) {upscale\\$\to$ 4K};
  \draw[flow] (asr) -- (llm);
  \draw[flow] (llm) -- (tts);
  \draw[flow] (tts) -- (av);
  \draw[flow] (av) -- (up);
\end{tikzpicture}
\caption{The fully streaming chain: each stage begins before the previous
finishes, so end-to-end \ttft{} approaches the slowest stage plus one first
frame. Upscaling is on-site only.}
\label{fig:stream}
\end{figure}
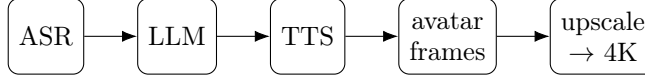

Third, \textbf{adaptive fillers and end-to-end streaming}: when synchronous
retrieval is unavoidable, a locally cached pool of Roosevelt-voiced acknowledgments
can begin immediately with no network round-trip; selection accounts for estimated
remaining latency, filler duration, and recent filler history, and the filler never
replaces the grounded answer path. Meanwhile, the whole chain streams end to end
(Fig.~\ref{fig:stream})---ASR $\rightarrow$ token-streaming LLM (with incremental
extraction of the spoken field from the JSON envelope) $\rightarrow$
sentence-chunked TTS $\rightarrow$ talking-head frames (and, on site, real-time
upscaling)---so end-to-end latency approaches the slowest single stage plus one
first frame rather than the sum of stages.

\paragraph{Measured latency in deployment.}
Over one exhibition period we logged 653 push-to-talk (PTT) releases. Of these, 457
($70\%$) ran to a completed answer; 156 were superseded by a newer press---a visitor
interrupting or re-asking---while two smaller groups produced an audible repeat
request (20) or a first-response timeout when a visitor kept the button held (20).
Across the 457 completed answers, the \emph{release-to-first-speech} delay---our
end-to-end \ttft{}, from PTT release to the first synthesized audio---averaged
$2.80$\,s (median $2.55$\,s, maximum $7.14$\,s), with $97\%$ of answers under $5$\,s
and $69\%$ under $3$\,s (Table~\ref{tab:latency}).

\begin{table}[t]
\centering
\begin{tabular}{lccccc}
\toprule
Release-to-first-speech & Mean & Median & Max & ${<}5$\,s & ${<}3$\,s \\
\midrule
completed answers ($N{=}457$) & $2.80$\,s & $2.55$\,s & $7.14$\,s & $97\%$ & $69\%$ \\
\bottomrule
\end{tabular}
\caption{End-to-end first-token latency measured in deployment---from push-to-talk
release to first synthesized speech---over 457 completed-answer turns (of 653 total
PTT releases; the remainder were interruptions, repeat requests, or user-hold
timeouts).}
\label{tab:latency}
\end{table}

Beyond these aggregates, two observations shaped the design. First, the
dominant latency cost is \emph{whether} a turn retrieves synchronously at all, rather
than the size of the index---which is why the self-routing and prefetching above
matter more than index engineering. Second, the length of the LLM context is itself a
latency lever, tying the context management of \S\ref{sec:context} to responsiveness
and not only to memory hygiene.

\subsection{Dynamic Context Management}
\label{sec:context}

The exhibit runs a single long-lived session throughout the day, so that visitors
incur no per-visit rebuild cost. Consequently, a single append-only conversation
history accumulates across many visitors; this history can neither fit within the
LLM context window nor be exposed to it across visitors. We therefore decouple the
\emph{transport} (session, sockets, queues), which persists all day, from the
\emph{context} actually supplied to the LLM, which is a sliding, resettable window.

Three mechanisms implement this. (1) \textbf{Watermark isolation}: each visitor
sees only history past a moving watermark advanced on entry, giving zero-copy
per-visitor isolation regardless of how the previous visit ended. (2)
\textbf{Bounded cross-phase bleed}: a short, time-bounded window of recent ambient
content is admitted so that a visitor who arrives mid-story can refer to it. (3)
\textbf{In-place refresh}: a watchdog periodically discards raw transcript and
transient state while \emph{preserving} only what is needed to keep the current
visit coherent (e.g., a per-visit summary) and without dismantling the session or
its transport, in the manner of summarize-and-recall memory designs
\citep{park2023generative,packer2023memgpt}. The system does not preserve
visitor-linked personal memory across separate visits; longer-lived records are
limited to non-identifying operational logs governed by the policy in
\S\ref{sec:epdg}. Because context length directly
affects \ttft{} (\S\ref{sec:latency}), this management is not only a memory-hygiene
measure but a latency lever. In our deployment the refresh fires when the microphone
has been idle for $\geq$180\,s, when the history exceeds $\approx$800 entries, or
when the session exceeds $\approx$8\,h; the per-turn context is assembled as the
current visitor's watermark slice plus a bounded ($\approx$90\,s) window of recent
ambient content.

\subsection{Safety for a Public Kiosk}
\label{sec:safety}

Because the exhibit is public and includes children, it must resist prompt
injection and steer away from improper content
\citep{greshake2023injection,rebedea2023nemo,inan2023llamaguard} under one strict
invariant: \emph{a safety check must never make the avatar stall or fall silent}.
In a face-to-face museum setting, a guardrail that blocks the utterance in progress
constitutes a failure of the interaction rather than a safeguard. We therefore pair
a three-layer, progressively heavier, and entirely non-blocking \emph{model-level}
guardrail stack (Table~\ref{tab:safety}) with a set of \emph{deployment-level}
interaction controls gated by curator review.

\paragraph{Three-layer guardrail stack.}
Each layer is heavier and later than the last, and none blocks the turn in flight.
\textbf{L1} screens every visitor input with regular expressions (prompt-injection
patterns and threat terms) synchronously in ${\sim}0$\,ms; on a hit it does not
terminate the session but injects an in-character ``deflect and pivot'' directive
into the \emph{next} turn, so the conversation continues in Roosevelt's voice.
\textbf{L2} runs a fast small-model classifier (\texttt{gpt-4.1-nano}) on the same
input \emph{asynchronously} and fire-and-forget; it is \emph{fail-open}, so if the
verdict has not returned by the time the reply is ready, the reply plays as normal
and any hit is handled on the following turn. \textbf{L3} reviews \emph{the avatar's
own outgoing line} with a mid-tier model (\texttt{gpt-4.1-mini}) \emph{in parallel
with the stream}, under a concurrency cap and skipping short boilerplate lines; it
defaults to an \texttt{observe} mode (watch, do not block) and can be switched to
revise-or-block when a deployment requires it.

\begin{table}[t]
\centering
\small
\begin{tabular}{@{}llll@{}}
\toprule
Layer & Target & Timing & Blocks the turn? \\
\midrule
L1 regex          & input  & sync, ${\sim}0$\,ms       & no --- directive next turn \\
L2 \texttt{nano}  & input  & async, fire-and-forget    & no --- fail-open \\
L3 \texttt{mini}  & output & parallel with the stream  & optional (default observe) \\
\bottomrule
\end{tabular}
\caption{The three-layer guardrail stack: progressively heavier, all non-blocking,
so that safety never stalls the avatar.}
\label{tab:safety}
\end{table}

\paragraph{Deployment-level interaction controls.}
Around the model, the deployed exhibit adds controls that bound the experience
without scripting the dialogue: pre-authored openings set
the interpretive frame, persona-boundary prompts constrain speculative drift, a
fallback catalog supplies a consistent degraded-mode response when retrieval is
weak or delayed (without substituting fabricated content), and an escalation policy
decides when to redirect or end a turn. Expert \emph{curator review} of grounding,
historical restraint, and boundary compliance gates public deployment
(\S\ref{sec:principles}).

\paragraph{Why fail-open.}
The stack is deliberately parallel, asynchronous, and fail-open: it tolerates a
detection that arrives one turn late in preference to any check that could block the
utterance being spoken. This choice is informed by an early failure mode, in which a
substring-based threat filter matched terms such as ``kill'' within benign phrases
(for example, ``killer view'') and terminated the session irrecoverably. The current
design instead deflects in character and restricts matching to word boundaries.

\subsection{Autonomous Operation}
\label{sec:autonomy}

\sysname{} runs all day without an operator, so it has to keep itself healthy and
recover on its own. Several watchdogs run continuously and watch the parts most
likely to fail---avatar generation, the dialogue service, the network transport, and
the rendering stack---and when one looks unhealthy the system reacts automatically:
it reconnects, restarts the service, or falls back to a safe degraded mode. Two
mechanisms illustrate the style. A keepalive pings the connection about every
30~seconds, so an idle link is not quietly dropped before the first visitor of the
day arrives; and a lightweight health probe sends a one-token request to the language
model, which both catches a hung connection and, by keeping it warm, helps prevent
the failure in the first place.

\begin{figure}[t]
\centering
\includegraphics[width=0.6\linewidth]{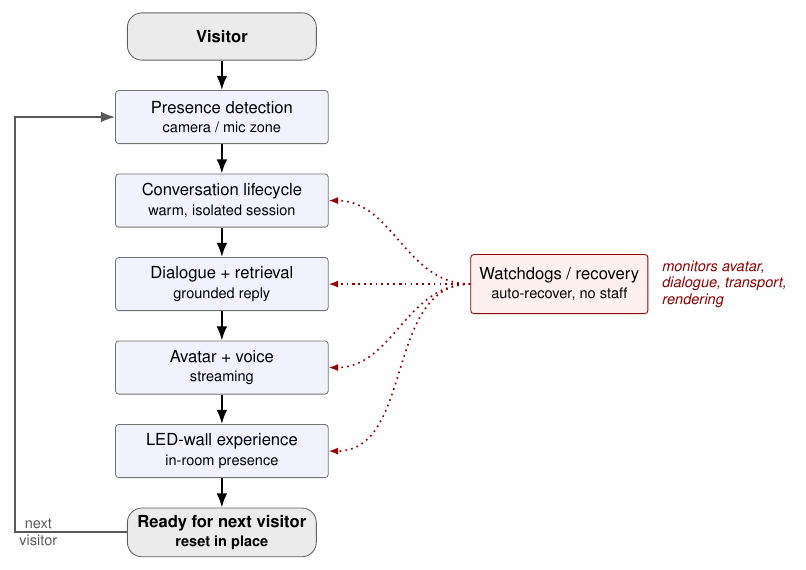}
\caption{The visitor lifecycle. Presence detection admits a visitor; the
conversation is managed as a warm, isolated session; responses stream through
dialogue, retrieval, avatar, and voice onto the LED-wall experience; watchdogs
recover any failed layer; and the session is reset in place, ready for the next
visitor---without restart or staff intervention.}
\label{fig:lifecycle}
\end{figure}

Between visitors, the system does not rebuild itself. A single session is kept warm
and reset in place (\S\ref{sec:context}), so each visitor gets the same fast,
consistent behavior as the last, with no restart. Together with per-visitor
isolation and services that can be restarted independently (\S\ref{sec:system}), this
lets one exhibit serve hundreds of visitors a day with little or no staff attention.
Figure~\ref{fig:lifecycle} follows a single visitor through this loop.

\section{Design Principles and Governance}
\label{sec:principles}

Building the Living Library at TRPL surfaced five recurring principles that
governed decisions across every layer, independent of the specific technology
choices described above. We state them explicitly here because they are what we
believe transfers to another institution's collection even where the technical
stack does not.

\subsection{Grounded in Truth}
Every user-facing output must be traceable to a verified source. In practice this
means the persona is grounded non-parametrically in primary sources rather
than baked into model weights or a static biography (\S\ref{sec:related}), \CEAG{}
answers modern questions through an attested historical analog rather than
by unconstrained generation (\S\ref{sec:ceag}), and records remain subject to continuous
expert verification and correction thereafter (\S\ref{sec:ocr}). Retrieval grounding is known to reduce conversational
hallucination in general \citep{shuster2021retrieval}; here it is also what makes
attributing a real historical figure's statements defensible.

\subsection{Human-in-the-Loop}
Automation accelerates the pipeline; it does not replace curatorial judgment at
the points that matter. Rather than hold the corpus behind an all-or-nothing
review, records are published to the index as they are processed, with their review
status and AI-generated provenance explicitly tracked. Domain experts subsequently
verify and correct records through the Archivist App (\S\ref{sec:ocr}). Curatorial
sign-off therefore governs not what enters the corpus but its quality over time. The escalation
policy and deployment-level interaction controls exist
precisely so that a human decision-maker, not the model alone, sets the boundaries
of the experience.

\subsection{Scalable by Design}
The architecture treats specific vendors and models as configuration rather than
foundation. The OCR stage is model-pluggable, so the production model (GPT-4.1) is
a swappable choice validated by evaluation, not a hard-wired dependency
(\S\ref{sec:ocr}); the three real-time services (vision, dialogue, rendering) are
loosely coupled and independently restartable (\S\ref{sec:system}); and persona
prompts are overridable at runtime through a keyed store, so tone and policy
changes do not require a redeployment (\S\ref{sec:persona}). This is what makes
the same architecture plausible for a different collection, a different
historical figure, or a newer model generation.

\subsection{Experience-First}
A grounded, low-latency answer is necessary but not sufficient. Deployment
observation showed that immersion depended as much on avatar realism, display
fidelity, lighting, audio, and physical staging as on the underlying language
model (\S\ref{sec:eval}), which is why \S\ref{sec:presence} treats embodied
presence as a first-class design concern rather than a presentation detail bolted
on afterward, and why the real-time budget of \S\ref{sec:latency} is enforced as
strictly as any correctness constraint.

\subsection{Ethically Responsible}
\label{sec:ethics}
Generating \emph{novel} utterances on behalf of a real historical figure risks
attributing statements to that person that they never made, a risk avoided by
recorded-replay systems \citep{artstein2014timeoffset}. We therefore treat
historical attribution as a first-class design constraint. Archival grounding and
the persona and chronology controls described in \S\ref{sec:persona}--\S\ref{sec:ceag}
are intended to keep generated responses close to the documentary record, and
visitors are informed that the responses are AI-generated.

Three residual concerns remain. First, archival bias: a corpus centered on one
figure's correspondence over-represents that figure's perspective and those of
their correspondents. Curatorial review and careful framing of contested history
can mitigate, but not eliminate, these omissions and biases. Second, the boundary
between inference and fabrication is inherently imperfect: even a source-grounded
analog remains a novel utterance that the historical figure never spoke. Third,
the exhibit processes presence signals and spoken input, requiring explicit
privacy and data-governance controls, described in \S\ref{sec:epdg}.

\section{Ethics, Privacy, and Data Governance}
\label{sec:epdg}
\sysname{} is designed for anonymous, walk-up use in a public library setting.
The deployment minimizes data collection, does not intentionally link
interactions to real-world identities, and applies defined retention and access
controls.

\paragraph{Notice and participation.}
On-site notice informs visitors that they are interacting with an AI-generated
historical persona, that camera-based presence sensing is used, that speech is
transcribed during an interaction, and that limited operational logs are
retained. Spoken participation is opt-in: pressing the push-to-talk control
initiates speech processing, and visitors may end the interaction at any time.

\paragraph{Children and vulnerable visitors.}
The exhibit is designed for a general audience and does not request names,
contact details, or other personal information. Any guardian or supervision
requirements for minors are governed by TRPL's on-site visitor policy and
signage.

\paragraph{Presence sensing and RFID.}
The vision subsystem estimates presence and interaction state; it does not
perform identity or facial recognition. Camera frames, images, face embeddings,
and biometric templates are not retained. The subsystem emits only a transient,
non-biometric tracking identifier with an approximately 30-second lifetime.
RFID is limited to staff control and exhibit-state transitions and is not used
to identify visitors or construct visitor profiles.

\paragraph{Audio and operational logs.}
Microphone audio is processed as a transient ASR stream and is not retained as an
audio recording. Operational logs are stored on a TRPL-controlled system and
include timestamps, randomized session identifiers, ASR transcripts, model inputs
and outputs, retrieval metadata, safety verdicts, and system-health and latency
measurements. The system does not create identity fields or intentionally
associate logs with real-world identities. Personal information voluntarily
spoken by a visitor may nevertheless appear in a transcript and is subject to
the same retention and deletion controls.

\paragraph{Retention, deletion, and access.}
Operational logs are retained for up to 30 days and then deleted under the
deployment's retention policy. Where a record can be identified from an
approximate interaction date and time, a visitor may request earlier deletion
via on-site staff. Access is restricted to authorized TRPL operators and project
maintainers for troubleshooting, safety auditing, and approved evaluation.

\paragraph{Ethics review.}
The results reported in this paper are limited to aggregate operational telemetry
and do not analyze or publish identifiable visitor interactions. Any research use
of visitor-level transcripts or behavioral data requires prior institutional
ethics review or a formal determination of exemption in accordance with the
applicable institutional policy.

\section{Replicability Framework}
\label{sec:replicability}

The preceding sections describe one deployment of the Living Library. Here we
distill the general process by which another institution could apply the same model to
its own holdings, with Layer 4 as an explicit, optional final step rather than an
assumed destination.

\paragraph{Step 1: Assess collection readiness.} Inventory assets across the
institution's own fragmented repositories and identify digitization gaps, in the
manner TRPL's own institutional survey did across more than forty repositories. This step is policy and inventory work, not
engineering, and its outcome determines the scope of everything after it.

\paragraph{Step 2: Build the corpus (Layer 1).} Digitize and aggregate materials
into institution-controlled storage, and establish data governance (source
identifiers, rights status, and a schema that separates immutable hard metadata
from later enrichment) before any AI processing begins (\S\ref{sec:layers12}).

\paragraph{Step 3: Apply AI processing (Layer 2).} Run OCR and metadata
enrichment as a model-pluggable stage, validate model choice with a
source-grounded evaluation framework on a stratified sample, and plan for expert curator review (\S\ref{sec:ocr}).

\paragraph{Step 4: Implement the retrieval system (Layer 3).} Index the corpus for hybrid dense/semantic search and expose it through a governed
contract; this step alone is sufficient to produce a researcher-facing tool in
the style of TRPL's Campfire (\S\ref{sec:layers12}), independent of whether a
conversational layer is ever added.

\paragraph{Step 5: Add a conversational layer.} Integrate an LLM against the
Layer 3 retrieval contract and define interaction patterns (persona grounding,
analogical reframing for out-of-scope queries, and a non-blocking safety
stack) following \S\ref{sec:persona}--\S\ref{sec:safety}.

\paragraph{Step 6 (optional): Avatar experience.} Where an institution wants an
embodied, real-time presence rather than a text or voice-only assistant, design a
persona and narrative catalog, and implement voice and visual rendering under the
same real-time budget and autonomous-operation discipline as \sysname{}
(\S\ref{sec:latency} to \S\ref{sec:autonomy}, \S\ref{sec:presence}).

Steps 1 to 4 are the load-bearing, broadly transferable core: they require
governance and engineering effort proportional to collection size but no
persona-design or real-time-systems expertise. Steps 5 and 6 are where an
institution takes on the harder, more open-ended problem this paper spends most
of its length on, and where the design principles of \S\ref{sec:principles}
matter most.

\section{Impact and Outcomes}
\label{sec:impact}

\subsection{For Institutions}
The immediate effect of Layers 1 to 3 is to unlock previously inaccessible or
underprocessed assets and make them consumable through a single governed
contract, rather than through per-repository, per-format access paths. Because
Layer 4 is optional and additive, an institution can realize this benefit without committing to a conversational-avatar
deployment at all, then extend to Layer 4 later against the same corpus.

\subsection{For Researchers and Historians}
A unified, semantically searchable index over a previously fragmented collection
shortens the path from question to relevant primary source, and supports queries
that cut across repositories and document types in ways manual finding aids do
not (\S\ref{sec:layer3}). We do not report a controlled time-to-insight study here;
this is a claim about what the retrieval layer structurally enables, not a
measured outcome, and we flag it as such in \S\ref{sec:limitations}.

\subsection{For the Public}
Layer 4, where built, turns a governed corpus into an encounter rather than a
search task. \S\ref{sec:eval} reports what we can measure from running \sysname{}
publicly (close to 5{,}000 visitors in its first two weeks) and what we
cannot (whether that encounter is faithful, free of anachronism, and experientially present) to the standard a dedicated study would require
(\S\ref{sec:limitations}). We report the deployment honestly rather than
overstate what two weeks of observational logging can establish.

\section{Limitations and Future Work}
\label{sec:limitations}

Our study concerns a single figure and a single archive; OCR errors are reduced by the expert review of \S\ref{sec:system}
but not eliminated, and residual errors may still propagate into retrieval and hence
into faithfulness; the on-site deployment exhibits higher latency than the web
deployment owing to the upscaling stage; and the self-routing mechanism may retrieve
when retrieval is unnecessary, or omit it when it is required. Our evidence is at
present largely observational: the faithfulness, experiential, and
autonomous-operation claims rest on deployment observation rather than controlled
measurement, and a larger visitor study and longitudinal reliability logging are
needed to substantiate them.
Finally, several components, such as the real-time avatar generation and the physical rendering
stack, are provided by partners and were tuned outside the core dialogue system.

\paragraph{Open questions we do not claim to settle.} This paper is a deployment
experience report, not a controlled evaluation, and we are explicit that three
qualities central to this class of system remain open. Whether grounding keeps
synthesis \emph{faithful} to the record, how often the figure lapses into
\emph{anachronism}, and how strong the visitor's sense of \emph{experiential
presence} is are questions a live exhibit raises but cannot answer on its own: the
first two would require systematic annotation of responses against their sources, and
the last a dedicated visitor study with established presence instruments
\citep{rashkin2021ais,ji2023hallucination,vomlehn2001exhibiting,hornecker2006learning,sylaiou2022virtual}.
We flag them as directions for future work rather than results we report here.

\paragraph{Open questions for the framework more broadly.} Beyond this single
deployment, generalizing the Living Library model to other institutions raises
questions this paper does not settle: how to standardize metadata across
institutions with different cataloging histories; how to ensure long-term model
accuracy and governance as underlying models change; what best practices should
govern representing historical figures responsibly across different subjects and
sensitivities; how to measure educational and engagement impact rather than infer
it from anecdote; and what interoperability standards would let Layer 3 corpora
from different institutions be queried together. We leave these as directions for
future, cross-institutional work.

\section{Conclusion}
\label{sec:conclusion}
We have presented the \emph{Living Library}, a replicable four-layer framework, including digitization
and corpus creation, AI-powered processing, retrieval and reasoning, and an
optional conversational interface. All of these combined turn fragmented archival collections
into governed, conversationally accessible knowledge systems, which we present here
instantiated at the Theodore Roosevelt Presidential Library. Layers 1 to 3 alone
already unify a $\sim$300{,}000-record collection into a
searchable corpus, reviewable through the \textit{Archivist App}, and consumed by a researcher-facing \textit{Campfire }experience. Layer 4 becomes \textit{\sysname{}}, a deployed conversational digital human
that embodies a real historical figure and grounds its responses in that same
corpus. Its design reconciles three competing requirements: faithfulness,
non-anachronism, and engagement under a real-time budget. The central technique,
\ceag{}, reframes contemporary questions as retrievals for historically attested
analogs, enabling faithful engagement on topics the archive never anticipated. A
latency-aware dual-path retrieval scheme and an end-to-end streaming pipeline keep
the interaction responsive. Dynamic context management sustains an all-day,
multi-visitor session under a bounded context. Lastly, a non-blocking, fail-open safety
stack keeps the exhibit publicly deployable. Beyond the conversation itself,
\textit{\sysname{}} is engineered for museum-scale autonomous operation: layered
watchdogs, session isolation, and automated lifecycle management sustain an
unattended, all-day exhibit across hundreds of visitors and for embodied
presence, delivering the figure as a full-scale digital human within a curated
physical environment. Our design principles and 
replicability framework are what we believe can transfer to another institution's
collection, independent of whether it builds a Layer 4 at all. We report the
Layer 4 deployment as an experience rather than a controlled study, and flag
faithfulness, anachronism, autonomous-operation reliability, and experiential
presence as open questions for future work. More broadly, the \textit{Living Library}
illustrates that turning an archival collection into a conversational knowledge
system is best engineered not as a single language interface but
as a layered, replicable system in which governed corpus construction, grounded
conversation, autonomous operation, and embodied presence are inseparable.

\section{Acknowledgments}

We would like to thank our partner teams across Microsoft, including Industry Solutions Engineering, the Foundry Agent Voice team, Microsoft Elevate, and CELA HQ, for their partnership, expertise, and support throughout this work.

We are grateful to AIS for their contributions to the development of the \textit{Archivist App} and to Valorem Reply for their work in building and advancing the \textit{Campfire} experience. 

We also thank Lemon Slice, Future of Storytelling (FoST) and the many partners, collaborators, and contributors who helped bring the \textit{Talk to TR} exhibit experience to life.

Finally, we extend our deepest thanks to the Theodore Roosevelt Presidential Library team for their vision, trust, and partnership throughout this work. We are especially grateful to the Theodore Roosevelt Center and Dickinson State University for preserving, curating, and providing access to the archival collections that served as the foundation for these experiences. Their commitment to making Theodore Roosevelt's historical record broadly accessible made this work possible.

\bibliographystyle{plainnat}   
\bibliography{references}

\end{document}